\documentclass[journal]{IEEEtran}
\usepackage{graphicx}
\usepackage{booktabs} 
\usepackage{amsmath,amssymb,amsfonts}
\usepackage{algorithmic}
\usepackage[linesnumbered, ruled]{algorithm2e}
\usepackage{siunitx}
\usepackage{pifont}
\usepackage{diagbox}
\usepackage{multirow}

\usepackage{tikz}
\usetikzlibrary{positioning}
\usepackage{cite}
\usepackage{enumitem}
\usepackage{makecell}
\usepackage{marvosym}

\usepackage{hyperref}

\newcommand*{\circled}[1]{\lower.7ex\hbox{\tikz\draw (0pt, 0pt)%
		circle (.4em) node {\makebox[1em][c]{\scriptsize #1}};}}

\usepackage{blkarray}

\ifCLASSINFOpdf
\else
\fi

\begin{document}

\title{
Lifelong Learning of Scheduling Rules with Surrogate-assisted Genetic Programming for Dynamic Flexible Job Shop Scheduling}

\author{
  \IEEEauthorblockN{
    Luyao Zhu$^{1}$, Fangfang Zhang$^{1}$, Yi Mei$^{1}$, Ke Chen$^{2}$\textsuperscript{\Letter}, Mengjie Zhang$^{1}$
  }
  
  \IEEEauthorblockA{
    $^{1}$ \textit{Centre for Data Science and Artificial Intelligence \& School of Engineering and Computer Science},\\
    \textit{Victoria University of Wellington}, Wellington, New Zealand \\
    \{luyao.zhu, fangfang.zhang, yi.mei, mengjie.zhang\} @ecs.vuw.ac.nz
  }
  
  \IEEEauthorblockA{
    $^{2}$ \textit{School of Electrical and Information Engineering}, \textit{Zhengzhou University}, Zhengzhou, China \\
    chenkezixf@zzu.edu.cn
  }
}

\maketitle

\begin{abstract}
Genetic programming (GP) has been widely used to learn scheduling rules for dynamic flexible job shop scheduling (DFJSS). However, trained rules tend to only generalise to tasks of similar scales and distributions, and when task characteristics change, learned rules often suffer significant performance degradation. Moreover, fine-tuning pre-trained rules on new tasks often leads to catastrophic forgetting, where prior knowledge is lost. This motivates the need for GP methods that can adapt to new tasks while preserving past knowledge, leading to a lifelong learning paradigm. However, lifelong learning in GP for DFJSS remains largely under-explored. To address this gap, this paper proposes a surrogate-assisted GP method for lifelong learning to handle sequential heterogeneous DFJSS tasks with job or batch arrivals. The framework includes three key components: an adaptive two-stage training strategy to balance plasticity and stability, a task-similarity-based weighting mechanism to regulate knowledge reuse, and an improved surrogate model with distance constraints to enhance estimation accuracy. Experimental results show that the proposed method effectively mitigates catastrophic forgetting while maintaining strong adaptability, outperforming several compared methods and demonstrating the potential of lifelong GP in dynamic scheduling environments.

\end{abstract}


\IEEEpeerreviewmaketitle

\section{Introduction}

\IEEEPARstart{J}{\lowercase{ob}} shop scheduling (JSS) is a classical yet critical combinatorial optimisation problem that addresses a wide range of complex and practical scheduling tasks \cite{AGNETIS2025367}, including computing resource allocation in cloud computing \cite{DBLP:conf/cec/BennettNZ14}, operation processing in manufacturing systems \cite{negahban2014simulation}, and truck routing in container port scheduling \cite{10478109}. Dynamic flexible job shop scheduling (DFJSS) extends the traditional JSS and more accurately reflects real-world production environments. In DFJSS, each operation of a job can be processed by multiple alternative machines. Consequently, two decisions must be made: \textit{operation sequencing}, which determines the processing order of operations, and \textit{machine assignment}, which selects an appropriate machine for a ready operation. Moreover, these decisions must be made under dynamic environments. In this paper, the considered dynamic events are random arrivals of jobs \cite{wang2019dynamic} or batches \cite{zhu2025scheduling}.

DFJSS is an NP-hard problem \cite{sotskov1995np}. Traditional approaches, such as \textit{exact optimisation methods} \cite{simon1988integer,chen1998improvement} and \textit{approximate solution methods} \cite{conroy1991handbook, pezzella2008genetic}, generally perform poorly in DFJSS due to their limited ability to handle dynamic events efficiently. In contrast, \textit{scheduling heuristics} \cite{muhlemann1982job}, such as dispatching rules \cite{li2020gep}, serve as priority functions that rank machines or operations at decision points to efficiently generate solutions for DFJSS. Specifically, in DFJSS, a scheduling heuristic typically consists of a \textit{sequencing rule} for operation sequencing for a particular machine and a \textit{routing rule} for machine assignment. However, manually designing scheduling heuristics typically requires domain expertise, prior experience, and a costly trial-and-error process. To address these challenges, genetic programming (GP) \cite{koza1990genetic} has been widely adopted to automatically learn effective scheduling rules for DFJSS. 

By training on a diverse set of DFJSS instances, it is possible that GP can learn high-quality scheduling rules that generalise well to DFJSS tasks with similar characteristics (e.g., distributions and problem scales) \cite{10065588}. However, when the distributional characteristics or scales of test instances differ substantially from those of the training instances, the performance of the learned rules tends to deteriorate significantly \cite{zhu2026investigation}. Moreover, in real-world manufacturing environments, problem characteristics frequently change over time. For example, a manufacturing system may continuously receive orders with varying batch sizes, job quantities, urgency levels, and processing requirements across different production stages. This motivates the need for GP algorithms to possess the capability to continuously adapt to new tasks, naturally leading to a lifelong/continual learning paradigm \cite{de2021continual,van2019three}. In lifelong learning, effectively coping with real-world dynamics requires an intelligent system to incrementally acquire, update, accumulate, and exploit knowledge throughout its operational lifetime, rather than relying on static models trained offline \cite{wang2024comprehensive}. Such systems should balance plasticity (i.e., continuously acquiring knowledge from new tasks) and stability (i.e., maintaining performance on previously encountered tasks). 

In lifelong learning, a straightforward and commonly used approach is to use the model trained on previous tasks as the initialisation and continuously train it on new tasks, which is referred to as \textit{fine-tuning} in neural solvers \cite{zhou2023towards}. In GP, fine-tuning corresponds to preserving some evolved scheduling rules from previous tasks as part of the initial population for new tasks and then continuing the evolutionary process \cite{kattan2023genetic}. However, such continual adaptation often results in catastrophic forgetting, where knowledge acquired from earlier tasks is changed or lost during training on new tasks, resulting in performance degradation when re-evaluated on previously encountered tasks. To mitigate catastrophic forgetting, many lifelong learning methods have been proposed. However, most of these methods are designed for neural networks and rely on gradient-based optimisation or parameter regularisation \cite{kirkpatrick2017overcoming,pei2025libog}, making them difficult to directly apply to GP or less effective in GP-based learning. Sim et al. \cite{sim2015lifelong} explored lifelong learning for bin packing using an immune-inspired GP hyper-heuristic framework that maintains a repertoire of heuristics through an immune-network mechanism. However, the evolved heuristics remain static once generated, and the framework does not explicitly address catastrophic forgetting. Kattan et al. \cite{kattan2023genetic} investigated lifelong learning with GP for symbolic regression and showed that preserving a small proportion of elites from previous tasks can improve performance. However, their approach largely resembles fine-tuning and does not explicitly examine the forgetting issue. As a result, lifelong learning in GP remains largely under-explored.





Since the evolutionary direction in GP is guided by individuals' fitness, changes in task characteristics lead to shifts in the fitness landscape, making fitness a key signal driving the search in lifelong learning scenarios. To mitigate forgetting, a natural approach is to define the fitness function by considering the performance of GP individuals across all previously seen tasks (similar to experience replay). In this way, individuals that perform well across multiple tasks can be identified, thereby encouraging the evolution of more general scheduling rules and improving knowledge retention. In practice, however, obtaining complete performance on past tasks is often infeasible. Directly evaluating scheduling rules on previous tasks is computationally expensive. To address this limitation, surrogate models can be used to predict the performance on the previous tasks. Among these, KNN-based surrogates with phenotype representations are commonly adopted due to their efficiency \cite{DBLP:journals/ec/HildebrandtB15}. However, these models typically provide only rough predictions because they ignore the distance between the target individual and its nearest samples. When an individual is far from existing surrogate samples, the estimated fitness may become unreliable. This problem becomes more pronounced in lifelong learning, where substantial differences between tasks can lead to highly dissimilar samples and increase the risk of prediction errors.

These limitations highlight the necessity of developing lifelong learning mechanisms for GP and improving surrogate accuracy in lifelong learning settings. To this end, we propose a novel method, i.e., lifelong learning with surrogate-assisted GP (LLSGP), for DFJSS to effectively acquire knowledge from new tasks while mitigating catastrophic forgetting. Specifically, the contributions of this paper are as follows:

\begin{enumerate}

\item A novel method LLSGP for DFJSS is developed to handle sequential scheduling tasks with varying distributions and scales. By integrating elite preservation and surrogate-based experience replay, this method mitigates forgetting and facilitates knowledge reuse across tasks.

\item An enhanced surrogate-based experience representation is developed for lifelong GP, where more diverse behavioural samples are incorporated to provide more reliable estimation of previous task knowledge.

\item An adaptive two-stage training strategy is proposed to utilise surrogate-based experience. It provides a parameter-free transition from current-task learning to past-task consolidation, achieving a competitive balance between stability and plasticity across different scenarios.

\item A fitness weighting mechanism based on task similarity is presented, which dynamically adjusts the contribution of historical tasks to balance the trade-off between retaining prior knowledge and adapting to new tasks.

\item Extensive experiments on DFJSS scenarios demonstrate that the framework achieves high average performance and significantly reduces performance degradation when switching between tasks. This work is one of the early attempts to explore lifelong learning in GP for dynamic job shop scheduling problems.

\end{enumerate}


\section{Background}

\subsection{Problem Formulation}

In DFJSS environments, a set of machines 
\(\mathcal{M} = \{M_1, M_2, ..., M_m\}\) 
is responsible for processing arrival units
\(\mathcal{U} = \{U_1, U_2, ..., U_n\}\), 
where each unit can be a single job or a batch of jobs. 
Each arrival unit \(U_i\) consists of one or more jobs 
\(\mathcal{J}_i = \{J_{i1}, J_{i2}, ..., J_{ik}\}\), 
which arrive dynamically at the shop floor. 
All jobs within an arrival unit are independent and can belong to different orders. 
The details of each job are unknown until the unit arrives at the shop floor. 
Each job \(J_{ij}\) consists of a predefined sequence of operations 
\(\mathcal{O}_{ij} = \{O_{ij1}, O_{ij2}, ..., O_{ij{l_{ij}}}\}\). 
In DFJSS, an operation \(O_{ijr}\) can be processed on multiple eligible machines.

\vspace{1mm}
\noindent
\textbf{Notation:}  
\begin{itemize}
    \item $\mathcal{J}$: the set of all arrived jobs, with cardinality $|\mathcal{J}|$.
    \item $d_{ij}, w_{ij}$: due date, and weight of job $J_{ij}$.
    \item \(r_{ij}\), \(C_{ij}\): release time, and completion time of job \(J_{ij}\).
\end{itemize}

\vspace{1mm}
\noindent
\textbf{Optimisation Objective:}  

The objective considered in this paper is to minimise the mean weighted tardiness ($WT_{\text{mean}}$).

\begin{equation}
WT_{\text{mean}} = \frac{1}{|\mathcal{J}|} \sum_{J_{ij} \in \mathcal{J}} w_{ij} \cdot \max \{0, C_{ij} - d_{ij} \}.
\end{equation}

\vspace{0mm}
\noindent

\textbf{Constraints:}

\begin{enumerate}

\item \textit{Precedence constraint:} An operation \(O_{ijr}\) cannot start until its preceding operation \(O_{ij(r-1)}\) is completed.

\item \textit{Exclusive assignment constraint:} Each operation \(O_{ijr}\) must be assigned to exactly one eligible machine.

\item \textit{Non-overlapping constraint:} A machine \(M_k\) can process only one operation at a time.

\item \textit{Non-interruptible constraint:} Once an operation \(O_{ijr}\) starts on a machine \(M_k\), it must run to completion without interruption.

\end{enumerate}

\subsection{GP for DFJSS}

GP has been widely used as a hyper-heuristic approach to automatically learn scheduling heuristics for combinatorial optimisation problems \cite{nguyen2011genetic,DBLP:journals/asc/DurasevicJK16}. Its effectiveness can be attributed to four main advantages. First, GP programs can act as priority functions to rank operations or machines efficiently, making them well-suited for dynamic scheduling problems. Second, GP does not require the optimal structure of heuristics to be known in advance, allowing the discovery of new structures. Third, the flexible representation of GP enables diverse heuristic behaviours to emerge from different genotypes, thereby enriching the genetic pool and supporting the evolution of effective programs. Finally, GP-evolved scheduling rules offer a potential advantage in interpretability over black-box models, which improves their practical applicability and acceptance among end-users.

\subsubsection{GP Process for DFJSS}

The GP procedure for DFJSS consists of two stages: training and test. During training, a population of GP individuals is evolved through repeated evaluation, parent selection, and genetic operators. Each individual represents a scheduling rule pair, whose fitness is assessed by applying it to DFJSS instances. After the stopping condition is reached, the best-performing rule pair is obtained. This rule pair is then applied to unseen test instances to generate the final scheduling results.

\begin{figure}[t]
    \centering 
    \includegraphics[width=0.4\textwidth]{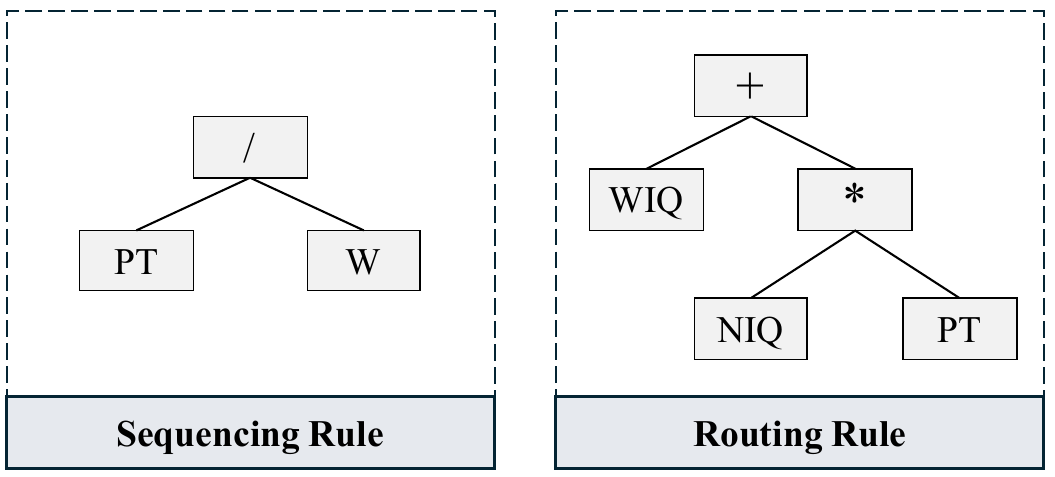}
    \vspace{-1mm}
    \caption{An example of GP-evolved scheduling rules.}
    \label{fig:GPInd}
\end{figure}

\subsubsection{Representation}

In this paper, a multi-tree representation is adopted to evolve scheduling heuristics for DFJSS \cite{zhang2018genetic}. Each GP individual comprises two trees (Fig.~\ref{fig:GPInd}): one encoding a sequencing rule and the other encoding a routing rule. The overall fitness of an individual is determined by the interaction between these two rules. The sequencing rule is defined by the priority function \(PT/W\), where \(PT\) represents the processing time of an operation on a specific machine, and \(W\) denotes the job weight. The routing rule is expressed as \(WIQ + (NIQ \times PT)\), where \(WIQ\) and \(NIQ\) correspond to the workload and the number of operations in a machine’s waiting queue, respectively. According to \cite{11043103}, these scheduling rules will select the candidate operation or machine with the lowest priority value.




\subsection{Related Work}

\subsubsection{Lifelong learning}

Lifelong learning, also referred to as continual learning \cite{wang2024comprehensive}, aims to enable models to sequentially learn multiple tasks and progressively accumulate knowledge in a manner analogous to human learning \cite{thrun1998lifelong}. Unlike multitask learning where all tasks are known beforehand, lifelong learning considers tasks that normally arrive sequentially over time, and task information is only available when the task appears. In addition, the key difference between lifelong learning and transfer learning is that lifelong learning considers not only the performance on the current (target) task, but also the performance on all previously encountered (source) tasks. Each task is typically associated with a distinct data distribution \cite{wang2024comprehensive}. A key challenge is maintaining a balance between plasticity and stability \cite{parisi2019continual}. Insufficient stability may lead to catastrophic forgetting, where performance on previously learned tasks degrades after learning new ones. Existing lifelong learning approaches are mainly designed for neural network based models and can be broadly categorised into three paradigms: parameter regularisation, model-based methods, and experience replay.

\textit{Parameter regularisation methods} attempt to preserve previously acquired knowledge by constraining updates to parameters that are considered important for earlier tasks. A representative example is Elastic Weight Consolidation \cite{kirkpatrick2017overcoming,pei2025libog}, which estimates the importance of each parameter using the Fisher information matrix and penalises large deviations of these parameters when learning new tasks. In this way, parameters critical for previous tasks are protected while allowing the model to adapt to new tasks. \textit{Model-based approaches} address catastrophic forgetting by allocating separate or partially shared models for different tasks \cite{wu2021incremental,ebrahimi2020adversarial}. For example, some methods maintain task-specific networks or dynamically expand the model structure to accommodate new tasks. However, these approaches typically assume that task identities are known and often require selecting the appropriate model when encountering a test task. As a result, they are less suitable for scenarios where tasks arrive sequentially and future task information is unavailable. Both parameter regularisation and model-based paradigms are not directly applicable to GP, since GP evolves symbolic program structures rather than continuous parameters, and the goal in GP-based scheduling is usually to evolve a single general scheduling rule instead of maintaining multiple task-specific models.

Among these paradigms, \textit{experience replay based methods} are more suitable for GP \cite{isele2018selective,buzzega2020dark}. These methods typically store experiences from previous tasks and replay them while learning new tasks to mitigate forgetting. However, although the underlying idea can be applied to GP, existing approaches may perform poorly in this context. For example, \cite{pei2025lifelong} proposed a lifelong learning method with behaviour consolidation (LLR-BC) for vehicle routing problems. Specifically, LLR-BC consolidates prior knowledge by incorporating the similarity between the decision-making behaviour of the current policy and that of the policy trained on previous tasks into the loss function. However, neural solvers typically rely on a single model, resulting in relatively stable decision-making behaviour. In contrast, GP is population-based, where multiple high-quality individuals may exhibit diverse decision-making behaviours. Therefore, behaviours that differ from the best-performing rule do not necessarily indicate poor performance.

\subsubsection{Lifelong Learning in Genetic Programming and Hyper-heuristic}

Lifelong learning can be integrated with various learning algorithms, including GP. However, research on incorporating lifelong learning mechanisms into GP remains relatively limited. \cite{azad2014simple} explored ``lifetime learning'' in GP, which differs from lifelong learning in that it focuses on the adaptation of a GP individual during its own ``lifetime'' (e.g., through local search), rather than the continuous acquisition and retention of knowledge across a sequence of tasks. Kattan et al.~\cite{kattan2023genetic} studied lifelong learning with GP for symbolic regression and found that preserving a small proportion of elite individuals from previous tasks (e.g., 5\% or 10\%) can achieve better performance than retaining the entire population. Nevertheless, this approach can be regarded as a form of fine-tuning and does not explicitly analyse or mitigate catastrophic forgetting. In parallel, lifelong learning have been incorporated into hyper-heuristic frameworks to select or generate heuristics across varied problem instances. Ortiz-Bayliss et al.~\cite{ortiz2015lifelong} proposed a lifelong learning selection hyper-heuristic for constraint satisfaction problems. Their approach employs a memory-based system of ``predictors'' and ``scenarios'' to store performance data for specific regions of the problem space. When a new task is encountered, the system either updates an existing predictor or creates a new one, thereby reducing interference with previously stored knowledge. Similarly, Sim et al.~\cite{sim2015lifelong} investigated lifelong learning in bin packing by proposing an immune-inspired hyper-heuristic framework that maintains a repertoire of heuristics through an immune-network mechanism. Hart et al.~\cite{hart2017towards,hart2022lifelong} later extended these ideas into a broader ``lifelong learning optimiser'' framework for continual optimisation. However, these approaches primarily rely on storing multiple specialised heuristics or predictors rather than evolving a single unified model that accumulates knowledge over time, which inherently limits their ability to capture cross-task generalizable patterns and requires substantial memory as the task sequence grows.

\subsubsection{Surrogate models of GP in DFJSS}

Surrogate models are widely used to approximate expensive real evaluations. In GP for DFJSS, two types of surrogate models are commonly adopted. The first evaluates GP individuals on simplified simulations with fewer jobs and machines \cite{nguyen2017surrogate}. However, this approach remains computationally expensive because DFJSS simulations still need to be executed. The second is KNN-based surrogate models with decision-making behaviours \cite{DBLP:journals/ec/HildebrandtB15}. In this model, each surrogate sample is represented by a decision-making behaviour vector and its corresponding fitness. For a candidate individual, the KNN model identifies the most similar sample (i.e., K=1) and predicts its fitness accordingly, making this approach computationally efficient. Therefore, KNN-based surrogate is adopted in this paper.

However, this surrogate generally provides only rough fitness estimates. In practice, it is often used as a pre-selection mechanism to identify promising individuals from a large candidate pool, after which the selected individuals are re-evaluated using real simulations. Zhu et al. \cite{zhu2023sample} reported an interesting exception: evaluating only one representative individual for each unique decision-making behaviour vector and assigning its fitness to identical individuals achieves performance comparable to evaluating the entire population. However, in most practical scenarios, the surrogate relies solely on the nearest neighbour and does not consider the actual distance to that neighbour. When the target individual is far from existing samples, the estimated fitness may become unreliable. This issue can be particularly problematic in lifelong learning settings, where tasks may differ substantially. In such cases, it may be difficult to find similar samples in the surrogate data accumulated from previous tasks, yet the surrogate still assigns the fitness of the nearest sample even when that sample lies far away in the behaviour space.

Moreover, surrogate fitness is typically derived from performance on a single instance, which may not adequately represent an individual’s performance over the entire task. These limitations suggest that the surrogate model needs to be improved to better support lifelong learning settings.

\begin{figure}[t]
    \centering 
    \includegraphics[width=0.5\textwidth]{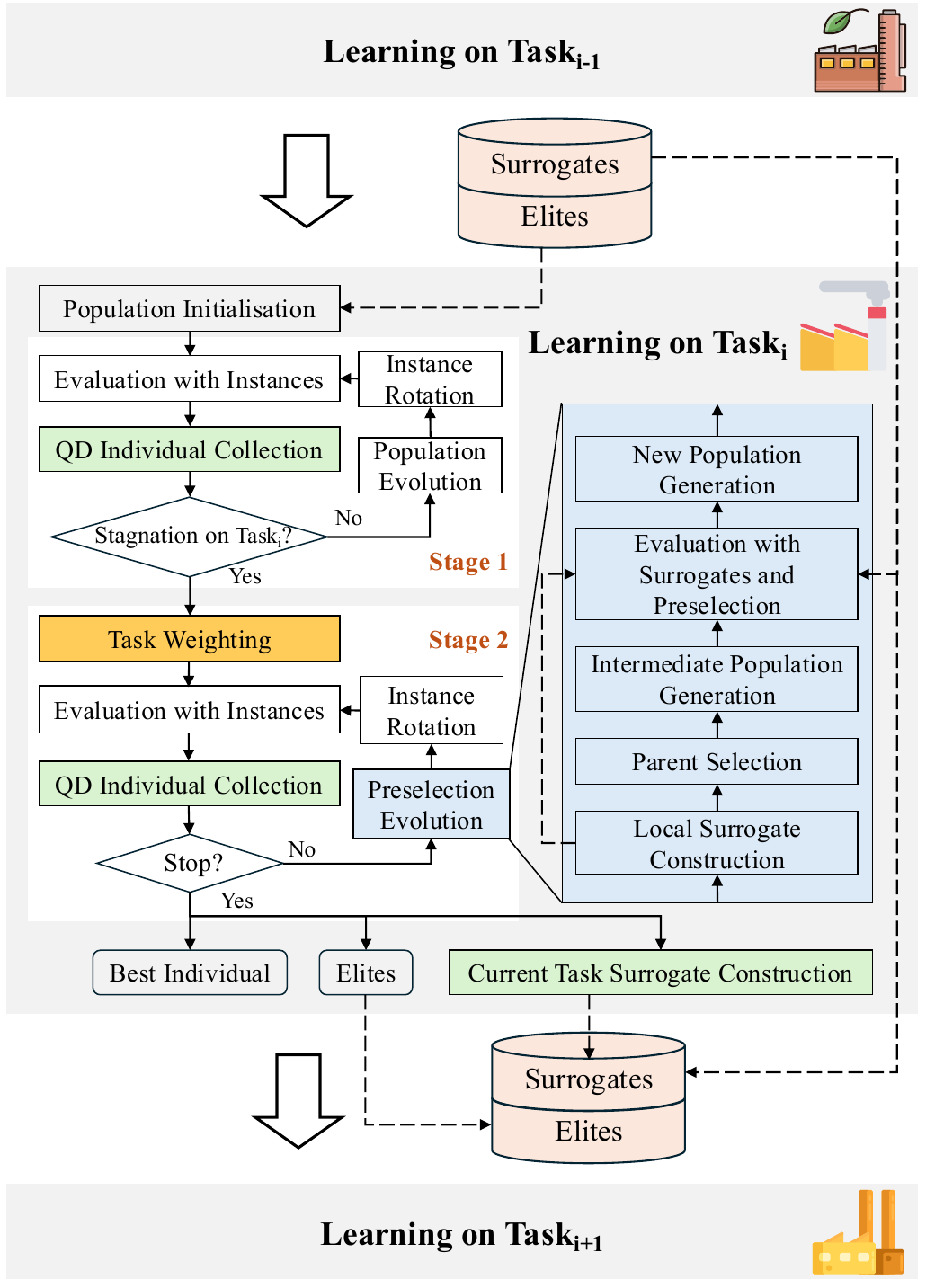}
    \vspace{-2mm}
    \caption{Process of LLSGP learning from sequentially arriving tasks.}
    \label{fig:flowChart}
\end{figure}

\section{Proposed Method}

\subsection{Overview of the Proposed LLSGP}

The main idea of LLSGP is to leverage fitness as a unified signal to guide the evolutionary search across tasks. Specifically, the fitness of each individual reflects both its performance on the current task and its estimated performance on previously seen tasks via surrogate models, thereby encouraging cross-task generalisation and mitigating catastrophic forgetting. Fig.~\ref{fig:flowChart} illustrates the flowchart of the proposed LLSGP. From top to bottom, LLSGP learns a sequence of tasks that arrive over time. When learning $task_{i}$, the algorithm reuses knowledge accumulated from previously learned tasks. The knowledge consists of elite individuals learned from the previous $task_{i-1}$ and surrogate models built from all previously seen tasks (i.e., $task_{1}$ to $task_{i-1}$). The overall learning process for $task_{i}$ is described as follows:

\begin{enumerate}

\item \textit{Population Initialisation:} 
The population is initialised by incorporating elite individuals transferred from the previous task. The remaining individuals are randomly generated to maintain population diversity. This initialisation strategy facilitates rapid adaptation to the new task while preserving diversity.

\item \textit{Adaptive Two-Stage Strategy:} 
To balance adaptation to the current task and retention of knowledge from previous tasks, an adaptive two-stage strategy is proposed. The Stage 1–2 transition is described in Section~\ref{Stage Switching Criterion}.

\item \textit{Stage 1:} This stage follows the standard GP process for learning a single DFJSS task. Each individual is evaluated on the training instances of $task_{i}$, enabling rapid adaptation to the current task. The population evolves through genetic operators (i.e., crossover, mutation, and reproduction) applied to selected parents. Training instances are rotated across generations to improve generalisation \cite{hildebrandt2010towards}. During this process, Quality-Diversity (QD) individuals are continuously collected to preserve diverse and high-quality individuals. When stagnation is detected on $task_{i}$, the algorithm switches to stage 2.

\item \textit{Stage 2:} Unlike Stage 1, which relies solely on current task performance, Stage 2 preselects the next population using a combined fitness metric that integrates both current performance and estimated historical fitness. A task-similarity strategy is first deployed to determine the weights for different task fitness. And then individuals are evaluated on the instances of $task_{i}$ to obtain the real fitness on the current task. During the preselection evolution phase, this process is enhanced using a surrogate-assisted preselection mechanism. Specifically, a local surrogate model is constructed using individuals from the current generation to estimate performance on the current task. Parents are then selected to generate an intermediate population whose size is $n$ times that of the standard population. To estimate general performance, all individuals within this intermediate population are evaluated by all built surrogates to obtain the estimated combined fitness. Finally, the top individuals, equal in number to the population size, are preselected based on this combined fitness to form the next generation. Consequently, the individuals within the new population tend to perform well on all seen tasks, thereby effectively mitigating the forgetting issue.

\item \textit{Termination and Knowledge Update:} 
If the termination criterion is met, the algorithm outputs the best evolved individual for test, along with a set of elite individuals for knowledge update. In addition, the collected QD individuals are used to construct a task-specific surrogate model, which is incorporated into the existing surrogate knowledge for future tasks. 

Notably, to ensure evaluation reliability, both parent selection and the final output of the best individual rely strictly on the real fitness of the current task, whereas the surrogate-estimated fitness is exclusively dedicated to preselecting individuals with good general performance.

\end{enumerate}

\subsection{QD-based Surrogate with Distance Constraints}

To reduce the computational cost of evaluating individuals across multiple tasks, a KNN-based surrogate model using decision-making behaviour vectors is employed to estimate fitness values on previously learned tasks \cite{DBLP:journals/ec/HildebrandtB15}.However, the estimated fitness of some individuals provided by the raw KNN-based surrogate model may be unreliable.

This limitation mainly arises from two aspects. First, surrogate samples are typically collected from individuals in the latest generation. 
Consequently, the collected samples often cover only a limited region of the behaviour space due to population convergence \cite{zhu2023sample}, which restricts the generalisation ability of the surrogate model. Second, traditional KNN-based surrogates (i.e., K=1) ignore the distance between samples. The fitness of the nearest neighbour is directly assigned to the candidate individual even when the distance is large. This issue becomes more severe in lifelong learning scenarios, where the behaviour distribution of individuals in new tasks may differ significantly from those observed in previous tasks.

To address these issues, an improved surrogate with QD-based sample collection and distance constraints is proposed.

\subsubsection{QD-based Sample Collection}

Instead of collecting samples from a single generation, surrogate samples are collected throughout the training process of the current task. Specifically, individuals that exhibit both good performance and behavioural diversity are recorded. To ensure the usefulness of the samples, only individuals whose fitness values outperform the reference dispatching rule, i.e., combining Weighted Shortest Processing Time (WSPT) and Workload in Queue (WIQ), are retained. This allows the collected samples to focus on promising regions of the behaviour space.

\begin{table}[t] 
	\centering
	\footnotesize 
	\setlength{\tabcolsep}{2mm}
	\caption{An example of the calculation of the behaviour vector of a sequencing rule across three decision points and each involving three candidate operations.}
	
	\label{tab:calculate PC} 
	\scalebox{1}
	{
		\newcommand{\tabincell}[2]{\begin{tabular}{@{}#1@{}}#2\end{tabular}}
		\begin{tabular}  { cccc} 
			\toprule 
			 	\footnotesize   {\tabincell{c}{Decision \\ Point}}   &  	\footnotesize {\tabincell{c}{Rank Obtained by \\ the Reference Rule}} & 	\footnotesize  {\tabincell{c}{Rank Obtained by \\ the Sequencing Rule}}  & 	\footnotesize $PC_i$ \\
			\midrule
			\(1(O_1)\) & {\underline{3}} & {\underline{1}} \\
			\(1(O_2)\) & 1 & 3 & {\textbf{3}}\\
			\(1(O_3)\) & 2 & 2 \\
			\midrule
			\(2(O_1)\) & 2 & 3 \\
			\(2(O_2)\) & {\underline{1}} & {\underline{1}}& {\textbf{1}}\\
			\(2(O_3)\) & 3 & 2 \\
			\midrule
			\(3(O_1)\) & 1 & 3 \\
			\(3(O_2)\) & 2 & 2 & {\textbf{3}}\\
			\(3(O_3)\) & {\underline{3}} & {\underline{1}} \\
			
			\bottomrule 
	\end{tabular}}
\end{table}

Behaviour diversity is measured using decision-making behaviour vectors, which are numeric vectors that contain the rankings of machines and operations in various routing and sequencing decision points. Table~\ref{tab:calculate PC} illustrates the vector calculation for a sequencing rule using three decision points, each comprising three candidate operations. For instance, in the first point, the examined sequencing rule assigns the highest priority to $O_1$. Consequently, the rank of $O_1$ under the reference rule (i.e., 3) is recorded as the vector value for this point. The remaining vector components are obtained following the same procedure. Finally, vector of the examined sequencing rule is $[3,1,3]$. The vector for a routing rule is determined similarly but using routing decision points. In this manner, each individual is mapped to a behavioural vector to reflect its behaviours for decision-making. Following the commonly used settings in \cite{zhang2021surrogate,11317967}, 20 routing decision points and 20 sequencing decision points are predefined from an independent simulation before the lifelong learning process. The same decision points are shared across all tasks to ensure a consistent behaviour representation. Each decision point contains 6 candidate choices. Therefore, each individual is represented by a 40-dimensional behaviour vector. The first 20 dimensions encode the selected choices at the sequencing decision points, while the remaining 20 dimensions encode those at the routing decision points. Each dimension takes an integer value from 1 to 6. Individuals with identical behaviour vectors are considered behaviourally similar; otherwise, they are regarded as behaviourally diverse.

After completing training on the current task, the collected individuals are evaluated on a validation set of 10 instances from the current task. The average fitness across instances is used as the final fitness value of each surrogate sample. Using multiple instances reduces evaluation noise and provides a more reliable estimation. To control the computational cost of building the surrogate, the number of surrogate samples $k$ is limited by $k = \min(P, N_c)$, where $P$ denotes the population size and $N_c$ denotes the number of collected individuals. If $N_c$ exceeds $P$, $k$-means clustering is applied to the behaviour vectors with $k = P$ to select more diverse samples. The cluster representatives are selected as the final surrogate samples, preserving behavioural diversity while keeping the sample size manageable. Since the arrival time of the next task is typically unknown in lifelong learning, there is often an idle period after training on the current task. The additional surrogate construction process can be performed during this period and therefore does not increase the effective training time.

\begin{figure}[t]
    \centering 
    \includegraphics[width=0.48\textwidth]{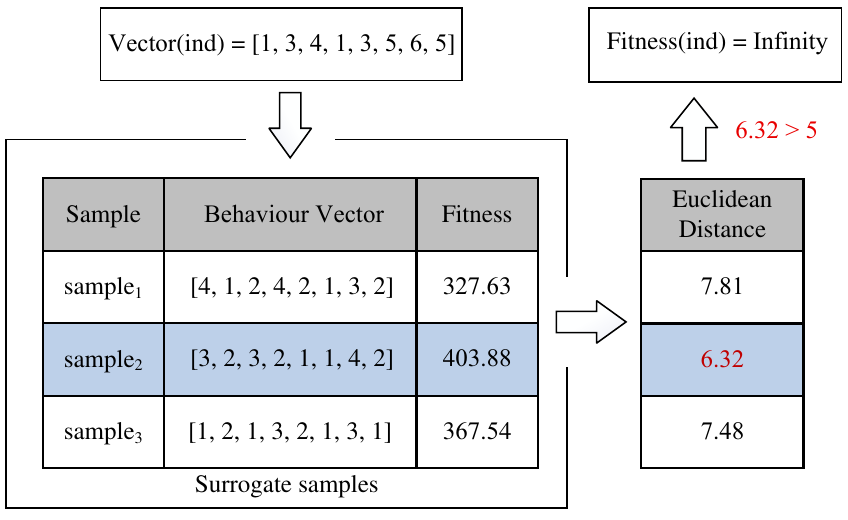}
    \vspace{-2mm}
    \caption{An example of distance-constrained fitness estimation.}
    \label{fig:surrogate}
\end{figure}

\subsubsection{Distance-Constrained Fitness Estimation}

Fig.~\ref{fig:surrogate} illustrates the distance-constrained surrogate-based fitness estimation. Each surrogate sample is represented by its decision-making behaviour vector and the corresponding average fitness \cite{DBLP:journals/ec/HildebrandtB15}. Specifically, Euclidean distance is adopted to measure phenotypic similarity between behaviour vectors, following established practice in surrogate-assisted GP for scheduling \cite{DBLP:journals/ec/HildebrandtB15, zhang2021surrogate}. Unlike Hamming distance, which disregards difference magnitudes, or rank-based metrics, which assume cross-situational rank orders, Euclidean distance emphasises larger behavioural deviations across decision situations.

For a candidate individual $\mathbf{ind}$ with vector $[1,3,4,1,3,5,6,5]$, the surrogate model first identifies its nearest neighbour in the surrogate sample set, i.e., $\mathbf{sample}_{2}$, based on the Euclidean distance of vectors. If the distance between $\mathbf{ind}$ and $\mathbf{sample}_{2}$ (i.e., $6.32$) is no greater than a predefined threshold $\tau$ (set to 5), the fitness of $\mathbf{sample}_{2}$ is used as the estimated fitness; otherwise, the estimation is considered unreliable and $\mathbf{ind}$ is assigned an infinite fitness value. This surrogate does not need to mathematically predict the complex formulation of objectives. In addition, note that since the dimensionality of the behaviour vector remains consistent across tasks, a fixed distance threshold can be applied across different tasks. This distance constraint prevents inaccurate surrogate predictions when the candidate individual is far from existing samples, which is particularly important in lifelong learning settings where behaviour distributions may change across tasks.

\subsection{Adaptive Two-stage Training Strategy}

To balance the optimisation of the current task and the reuse of knowledge from previously learned tasks to mitigate forgetting, an adaptive two-stage training strategy is adopted.

\subsubsection{Stage 1}

In Stage~1, the algorithm focuses solely on the current task. Accordingly, an individual's fitness is defined by its performance on the current task, without considering previously seen tasks. This allows the search to adapt freely to the new environment and improves plasticity. Since the initial population includes elite individuals transferred from the previous task, the search naturally starts from GP individuals that are effective in earlier tasks. Stage~1 therefore mainly performs local adaptation around these transferred elites, adjusting them to the new task and generating individuals for the current task. Although this process may degrade performance on previous tasks, it provides a strong starting point for discovering high-quality GP individuals for the current task.

\subsubsection{Stage Switching Criterion}
\label{Stage Switching Criterion}

The transition from Stage~1 to Stage~2 is determined using a detection mechanism based on population improvement. Specifically, the mean normalised fitness of the population on the current task is recorded at each generation. Since the evaluation instances vary across generations, all fitness values are dynamic-scaled via min--max normalisation, where the minimum and maximum bounds are the best fitness within the current population and the fitness of a reference dispatching rule combining WSPT and WIQ, respectively. The mean normalised fitness reflects the global convergence state and overall evolutionary potential of the entire population. Since the lower bound (0) always represents the current best individual, a population mean closer to 0 indicates that most individuals have clustered around the optimal solution, meaning the local search has reached its limit. Conversely, a larger mean value implies that the population still maintains diversity and room for further improvement. Consequently, this population-level normalisation metric remains stable across generations. To detect whether further improvement on the current task has become limited, a sliding window of size $W$ is employed to track the mean normalised fitness from the most recent generations. If the average historical improvement within this window falls below a predefined threshold $\epsilon$, the optimisation progress is deemed to have stagnated, thereby triggering the switch to Stage~2. This switching condition is mathematically formalised below.

Let $f_t$ denote the mean normalised fitness of the population at generation $t$. The average historical improvement over a sliding window of size $W$ at generation $t$ ($t > W$) is formally defined as:
    \begin{equation}
        \Delta_t = \frac{1}{W-1} \sum_{k=t-W+1}^{t-1} (f_k - f_{k+1})
    \end{equation}
    The transition from Stage 1 to Stage 2 is triggered at the earliest generation $t^*$ that satisfies the following condition:
    \begin{equation}
        t^* = \min \left\{ t \;\middle|\; \Delta_t \le \epsilon \right\}
    \end{equation}

This mechanism allows the algorithm to first rely on real fitness evaluations to guide the search toward promising regions of the current task. When further improvement becomes limited, knowledge from previously learned tasks is introduced to enhance search stability while maintaining the plasticity established in Stage~1.

\begin{table*}[t]
    \centering 
    \caption{Parameters of Four Designed DFJSS tasks.}
    \label{tab:4workshops}
    \footnotesize
    \begin{tabular}{ccccccc}
        \toprule 
        \bfseries Task & \bfseries Machines & \bfseries Jobs & \bfseries Utilisation & \bfseries Due Date Factor & \bfseries Batch Size & \bfseries Distribution\\
        \midrule 
        
        Standard Discrete Job Shop with Job Arrivals \cite{10065588}  
        & 10 & 3000 & 0.85 & 1.5 & [1,1] & Uniform$[1,99]$\\
        
        Standard Discrete Job Shop with Batch Arrivals \cite{zhu2025scheduling}  
        & 10 & 3000 & 0.85 & 2.0 & [10,20] & Exponential($\mu=20$)\\
        
        Low-Demand Small-Order Job Shop \cite{stevenson2005review}  
        & 2 & 500 & 0.60 & 1.2 & [1,9] & Gamma($k=2$, $\theta=25$)\\
        
        Heavy-Demand Large-Order Job Shop \cite{stevenson2005review}   
        & 15 & 3000 & 0.95 & 4.0 & [20,30] & Normal($\mu=80$, $\sigma=10$)\\

        \bottomrule
    \end{tabular}
    \footnotesize
    \textit{Note}: All processing times are truncated to the range $[1,99]$.
\end{table*}

\subsubsection{Stage 2}

The fundamental distinction between the two stages lies in their respective evolutionary procedures. Stage~1 focuses exclusively on the current task by directly generating a standard population of size $N$ for the next generation. In contrast, Stage~2 incorporates a global multi-task perspective by introducing an intermediate population of size $m \times N$, from which $N$ individuals are preselected based on their estimated performance across all seen tasks. Specifically, the fitness of each individual within this intermediate population is predicted entirely using the respective surrogate models of all tasks, without involving any true simulation evaluations. An individual's combined fitness is defined as a weighted combination of these estimated values (weight is described in the following section), and the top $N$ individuals are retained to form the next generation. In this manner, the newly formed population tends to perform well across all encountered tasks, thereby naturally mitigating catastrophic forgetting.

To remove the estimated fitness difference across different tasks, all estimated fitness values are changed using min--max method, where the minimum and maximum values are fixed to the best recorded fitness within the surrogate samples and the fitness of a reference rule (combining WSPT and WIQ), respectively.
In addition, introducing this intermediate population offers another critical advantage. Due to potential dissimilarity between tasks, many newly generated individuals may lie far from surrogate samples constructed from previous tasks. In the distance-constrained surrogate model, such individuals are assigned infinite fitness, indicating unreliable estimation. Retaining these individuals in the population could waste computational resources on poorly performing solutions. The proposed preselection strategy naturally discards these unreliable individuals during the filtering process, thereby ensuring that only high-performing candidates advance.

Overall, the proposed adaptive two-stage strategy first exploits real fitness evaluations to guide the search towards promising regions of the current task. Once the optimisation progress becomes limited, knowledge from previously learned tasks is introduced to improve search stability while preserving the plasticity established in the early stage.

\subsection{Task Weighting based on Task Similarity}

In Stage~2, the fitness of individuals is defined as a weighted combination of their performance across all previously seen tasks. During the learning of the current task, the degree of forgetting may vary across previously learned tasks: tasks that are more similar to the current task tend to experience less forgetting, while dissimilar tasks are more likely to suffer stronger forgetting. Therefore, the fitness contributions from different tasks should be assigned different weights. In addition, to balance plasticity and stability, both the weight of the current task and the total weight assigned to previous tasks need to be carefully considered. To this end, a task-weighting strategy based on task similarity is proposed.

The task similarity is estimated using the individuals in the final population of Stage~1. As these individuals have already adapted well to the current task through real fitness evaluations, their performance across tasks provides useful information for identifying task relationships. The individuals with duplicated behaviours are removed so that only one representative individual is retained for each unique behaviour vector. In addition, individuals with infinite fitness on previous tasks are also removed. Let $\mathcal{P}$ denote the remaining set of individuals. For each individual $k \in \mathcal{P}$, $F_i^k$ denotes its fitness on the current task $task_i$, and $F_j^k$ denotes its estimated fitness on a previous task $task_j$. The similarity between $task_i$ and $task_j$ is measured using the Spearman rank correlation coefficient

\begin{equation}
\rho_{i,j} = \mathrm{Spearman}(F_i, F_j)
\end{equation}
where $F_i = \{F_i^k \mid k \in \mathcal{P}\}$ and $F_j = \{F_j^k \mid k \in \mathcal{P}\}$.

The computed correlations are used to determine task weights when combining fitness values across tasks. The current task is assigned a fixed weight of 1 to ensure optimisation focuses on it, while the remaining tasks share a total weight of 1 based on their similarity to the current task, balancing plasticity and stability. Tasks with higher correlation receive smaller weights to reduce redundancy from highly similar tasks. The weight of a previous task $task_j$ is defined as

\begin{equation}
w_j = \frac{1 - \rho_{i,j}}{\sum_{k=1}^{i-1} (1 - \rho_{i,k})}
\end{equation}

Using these weights, the combined fitness of an individual is calculated as

\begin{equation}
F_{\text{combined}} = F_i + \sum_{j=1}^{i-1} w_j F_j
\end{equation}

where $F_i$ denotes the fitness on the current task and $F_j$ represents the estimated fitness on the previous tasks. The task weights are computed once after Stage~1 and remain fixed during Stage~2. This strategy allows the algorithm to emphasise tasks that are more likely to suffer forgetting while reducing redundant influence from highly similar tasks.

		


\subsection{Summary}

The proposed framework enable GP to effectively learn from a sequence of tasks in a lifelong learning manner, by reusing effective knowledge across sequential DFJSS tasks. The adaptive two-stage strategy first relies on real fitness to guide the search on the current task and then incorporates knowledge from previously learned tasks. The task-weighting mechanism further mitigates forgetting by adjusting the contribution of previous tasks according to their similarity with the current task. Meanwhile, the QD-based surrogate with distance constraints improves the reliability of surrogate-assisted fitness estimation. Overall, these components form an effective evolutionary framework for solving DFJSS under lifelong learning scenarios.

\section{Experiment Design}

\subsection{DFJSS Task Design}

The simulation technique is adopted to model DFJSS instances. In the simulation, $m$ machines process $n$ jobs. The number of operations per job and the number of candidate machines per operation are sampled from the range $[1, m]$ following uniform distributions. The processing time of each operation is independently sampled from $[1, 99]$. Following common settings in the literature \cite{zhang2020FS}, the processing times follow different distributions across tasks while remaining within the same value range. To model machine heterogeneity, the actual processing time on a candidate machine follows a uniform distribution within $[\mu - 10, \mu + 10]$, where $\mu$ is the baseline processing time. Job (or batch) arrivals follow a Poisson process, where the inter-arrival times are exponentially distributed. The mean inter-arrival time is adjusted according to the desired system utilisation level so that the workload of the system remains stable. Each arrival may contain a batch of jobs, and the batch size follows a uniform distribution within a predefined range. Job importance is modelled using weights: 20\% of jobs have weight 1, 60\% have weight 2, and 20\% have weight 4 \cite{DBLP:journals/ec/HildebrandtB15}. The due date of each job is calculated as its estimated processing time multiplied by a predefined due-date factor plus its arrival time.

To represent diverse and practically relevant manufacturing environments, four DFJSS tasks with different problem scales, workshop parameters, and processing time distributions are constructed in this study. More details are provided in Table~\ref{tab:4workshops}. Note that the warm-up period is set to 20\% of the target number of processed jobs. For example, in a task requiring 3,000 completed jobs for objective calculation, the first 600 jobs serve as the warm-up period, making the total simulation length 3,600 jobs.

For each task, separate training and test instance sets are constructed. In each generation, a distinct training instance from the current task is used \cite{hildebrandt2010towards}, while 50 unseen test instances are used to evaluate test performance \cite{10044237,tran2025online}. The model is trained sequentially across tasks following a predefined training order. After each task, it is evaluated on the test sets of all encountered tasks to assess both performance on the current task and the degree of forgetting on previous tasks. In addition, the source code is available at \url{https://anonymous.4open.science/r/Lifelong-learning-with-Surrogate-assisted-GP-for-DFJSS-CD77/}.

\subsection{Design of Comparisons}

Six random training orders are used to avoid the influence of task order. In the following, each task is represented by its job processing time distribution, i.e., \textit{E} (Exponential), \textit{G} (Gamma), \textit{U} (Uniform), and \textit{N} (Normal). The six training orders are EGUN, GENU, NUGE, UGEN, EUNG, and NEUG. The compared algorithms are described as follows.

\begin{enumerate}
    \item \textbf{Restart}: When switching to a new task, a new population is randomly initialised without using any knowledge from previous tasks.
    
    \item \textbf{Fine-tuning} \cite{kattan2023genetic}: When switching to a new task, a portion of high-quality individuals from the previous task are preserved to initialize the population for the new task, while the remaining individuals are randomly generated. Different retention ratios are considered, including 5\%, 10\%, and 100\%, referred to as Fine-tuning-5\%, Fine-tuning-10\%, and Fine-tuning-100\%, respectively.
    
    \item \textbf{LLR-BC} \cite{pei2025lifelong}: A state-of-the-art lifelong learning method based on neural solvers for vehicle routing problems. We adopt its idea of measuring the similarity of decision-making behaviours between candidate individuals and the best individuals from previous tasks. The similarity is incorporated into the fitness function, where individuals with behaviours more similar to the previous best individuals are assigned better fitness. This encourages the search to discover individuals that perform well on the current task while maintaining behavioural similarity with high-quality individuals from previous tasks.
    
    \item \textbf{LLSGP}: our proposed method.

    \item \textbf{Multi-task}: The joint multi-task baseline that trains a single population concurrently on all tasks. To ensure a fair comparison, the total number of real evaluations allocated to each task is kept exactly the same as the other sequential algorithms (i.e., $100$ generations with one instance per task in each generation). Its performance serves as the ideal upper bound for normalising the results across different tasks.
\end{enumerate}

		
        
		

\begin{table}[t]
	\setlength{\tabcolsep}{1.4mm}
	\centering
	\caption{The parameter settings in GP.}	\label{tab2} 
	\footnotesize
	\label{tab:parameter}
	\begin{tabular}  {c c}
		\toprule 
		\bfseries Parameter  & \bfseries Value\\
		\midrule 
	    Number of tasks & 4 \\
        Number of generations per task & 100\\
        Population size & 500\\
		
		Parent selection & Tournament selection \\
		&with size 7 \\
        Number of elites & 10\\
		\midrule
		
		Initial minimum / maximum depth &  2 / 6\\ 
		Maximal depth of programs & 8\\
		Crossover / Mutation / Reproduction rate & 80\% / 15\% / 5\% \\
        Terminal / non-terminal selection rate & 10\% / 90\% \\
        Method for initialising population  &  ramped-half-and-half \\ 
		\bottomrule
	\end{tabular}
\end{table}

Furthermore, to evaluate the effectiveness of the strategies embedded in LLSGP, several ablation variants are examined:

\begin{enumerate}
    \item \textbf{One-stage}: Remove the adaptive two-stage strategy and use the Stage~2 fitness definition throughout the entire evolutionary process.
    
    \item \textbf{Raw-surrogate}: Use the surrogate model without the proposed sample collection and distance constraints.
    
    \item \textbf{Equal-weight}: Assign equal weights to all tasks when computing the combined fitness.
    
    \item \textbf{Focused-weight}: Assign a weight of 1 to the current task, while the remaining weight is evenly distributed among previous tasks.
\end{enumerate}

\begin{table*}[t]
  \centering
  \footnotesize
  \caption{Comparisons of mean (std) metrics and ranks for the compared algorithms over six training orders (30 runs each).}
  \label{tab:results}
  \resizebox{1\textwidth}{!}{
  \begin{tabular}{l@{\hspace{0.1pt}}l cccccc}
    \toprule
    \bfseries Metric & & Restart & Fine-tuning-5\% & Fine-tuning-10\% & Fine-tuning-100\% & LLR-BC & LLSGP \\
    \midrule
    \multirow{2}{*}{AP}
      & Value
      & 69.92 (20.56) ($\uparrow$, 0.00)
      & 34.75 (7.72) ($\uparrow$, 0.01)
      & 43.05 (23.22) ($\uparrow$, 0.01)
      & 39.36 (15.37) ($\uparrow$, 0.00)
      & 34.08 (14.51) ($\uparrow$, 0.03)
      & \textbf{29.75 (12.15)} \\
      & Rank
      & 5.83 & 2.83 & 4.17 & 3.50 & 2.83 & \textbf{1.83} \\
    \midrule
    \multirow{2}{*}{AF}
      & Value
      & 86.68 (27.66) ($\uparrow$, 0.00)
      & 40.69 (10.34) ($\uparrow$, 0.00)
      & 51.66 (31.38) ($\uparrow$, 0.00)
      & 40.69 (10.34) ($\uparrow$, 0.00)
      & \textbf{32.68 (18.97)}($\approx$, 0.84)
      & 34.35 (15.88) \\
      & Rank
      & 5.83 & 3.00 & 4.33 & 3.67 & \textbf{1.83} & 2.33 \\
    \midrule
    \multirow{2}{*}{AMF}
      & Value
      & 131.21 (24.69) ($\uparrow$, 0.00)
      & 63.26 (15.75) ($\uparrow$, 0.02)
      & 68.81 (34.10) ($\uparrow$, 0.00)
      & 73.68 (25.61) ($\uparrow$, 0.00)
      & \textbf{49.11 (25.64)} ($\approx$, 0.54)
      & 53.57 (27.70) \\
      & Rank
      & 6.00 & 3.17 & 3.33 & 4.33 & \textbf{1.83} & 2.33 \\
    \midrule
    \multirow{2}{*}{APl}
      & Value
      & 5.02 (0.94) ($\approx$, 0.67)
      & 4.38 (0.77) ($\approx$, 0.15)
      & 5.24 (2.82) ($\approx$, 0.57)
      & 4.89 (0.82) ($\approx$, 0.84)
      & 10.17 (2.75) ($\uparrow$, 0.03)
      & \textbf{4.12 (1.26)} \\
      & Rank
      & 3.50 & \textbf{2.50} & 2.83 & 3.83 & 5.83 & \textbf{2.50} \\
    \midrule
    \multicolumn{2}{l}{Average Rank} & 5.29 & 2.88 & 3.67 & 3.83 & 3.08 & \textbf{2.25} \\
    \bottomrule
  \end{tabular}}
    \footnotesize
  \parbox{0.9\textwidth}{
\raggedright
  \textit{Note}: Bold values indicate the best mean performance for each metric. The arrows ($\uparrow$, $\downarrow$, $\approx$) denote that the proposed LLSGP performs significantly better, worse, or statistically similar to the compared algorithm, respectively. The number following the arrow is the associated $p$-value. The Average Rank represents the average rank of each algorithm across all metrics and training orders.}
\end{table*}

\subsection{Parameter Settings}

Each GP individual consists of terminals and functions. The terminals represent features of the job shop related to machines, operations, and jobs/batches, consistent with those used in \cite{zhu2025scheduling}.
The function set is defined as \{\(+\), \(-\), \(*\), protected \(/\), \(max\), \(min\)\}. The protected division returns one when the denominator is zero. The \(min\) and \(max\) functions return the minimum and maximum of their arguments, respectively. Other GP parameter settings are listed in Table~\ref{tab:parameter}. In addition, based on comprehensive parameter sensitivity analyses in supplementary materials, the window size $W$, the threshold $\epsilon$, and the distance threshold $\tau$ in the surrogate model are set to 5, 0.1\%, and 5, respectively. The size of the intermediate population in Stage 2 is set to $n=3$. This configuration also aligns with \cite{DBLP:journals/ec/HildebrandtB15}, providing a better balance between efficiency and effectiveness.

\subsection{Evaluation Metrics}

To comprehensively evaluate the performance of methods in a lifelong learning scenario, we define the following evaluation metrics, following standard practices in \cite{wang2024comprehensive}. Let \( d_{i,j} \) denote the normalised test performance, i.e., average objective value (mean-weighted-tardiness) over all the test instances of the learned scheduling rule on task \( T_j \), after sequentially training on the first \( i \) tasks (including \( T_j \)). Since performance scales differ significantly across tasks, all raw test results are normalised against the joint multi-task model to represent the relative performance gap. Specifically, the normalised performance \( d_{i,j} \) is calculated as follows:
\begin{equation}
d_{i,j} = \frac{F_{i,j}}{F^{\text{mt}}_j} - 1
\end{equation}
where \( F_{i,j} \) represents the raw average objective value over all test instances of task \( T_j \) after training up to task \( i \), and \( F^{\text{mt}}_j \) denotes the performance of the joint multi-task model on task \( T_j \). A value closer to 0 indicates that the algorithm's performance is closer to the ideal upper bound. To improve readability, the normalised values are multiplied by 100. For example, \( d_{3,1} \) represents the performance on task \( T_1 \) by the learned scheduling rule after training on tasks \( T_1 \), \( T_2 \), and \( T_3 \). Based on this notation, after learning \( k \) tasks, we calculate:

\begin{enumerate}
    \item \textbf{Average Performance (AP):} the average performance on tasks learned so far, reflecting the overall performance after learning all tasks, i.e.,
    \begin{equation}
    AP = \frac{1}{k} \sum_{i=1}^{k} d_{k,i}.
    \end{equation}

    \item \textbf{Average Forgetting (AF):} the average performance drop on previously learned tasks after training on all tasks, measuring the average forgetting on previously learned tasks, i.e.,
    \begin{equation}
    AF = \frac{1}{k-1} \sum_{i=1}^{k-1} \max(0, d_{k,i} - d_{i,i}).
    \end{equation}

    \item \textbf{Average Maximum Forgetting (AMF):} the average of the maximum forgetting experienced by each task during lifelong learning, capturing the worst-case forgetting during the learning process, i.e., 
    \begin{equation}
    AMF = \frac{1}{k-1} \sum_{i=1}^{k-1} \max_{j=i+1}^{k} \max(0, d_{j,i} - d_{i,i}).
    \end{equation}

    \item \textbf{Average Plasticity (APl):} the average performance on each newly encountered task immediately after training on it, indicating the ability to adapt to newly encountered tasks, i.e.,
    \begin{equation}
    APl = \frac{1}{k} \sum_{i=1}^{k} d_{i,i}.
    \end{equation}
\end{enumerate}

Smaller values indicate better performance for all metrics.

\section{Results and Discussions}

Each training order of each algorithm is independently run 30 times to assess performance and robustness. The results are analysed using the Wilcoxon rank-sum test and Friedman's test at a significance level of 0.05. 

\subsection{Overall Performance}
Table~\ref{tab:results} shows the average metrics over the six training orders. We first focus on AP, the primary metric that measures the overall performance of the learned scheduling rules across all seen tasks. The proposed LLSGP achieves the best AP value of 29.75, which is clearly lower than those of Restart (69.92), Fine-tuning-5\% (34.75), Fine-tuning-10\% (43.05), and Fine-tuning-100\% (39.36). LLR-BC also improves over the fine-tuning baselines and achieves a slightly higher AP value of 34.08. By selecting individuals according to behavioural similarity with the best rules from previous tasks, LLR-BC encourages the reuse of useful behaviours and therefore helps reduce catastrophic forgetting. However, this strategy mainly focuses on maximising similarity to previously best individuals. As a result, individuals that exhibit different but potentially effective behaviours may be overlooked during the selection process. In contrast, LLSGP directly considers individuals' performance across all previously seen tasks, encouraging the selection of individuals that perform well on multiple tasks, which leads to better overall performance.

\begin{figure}[t]
	\centering 
	\includegraphics[width=0.47\textwidth]{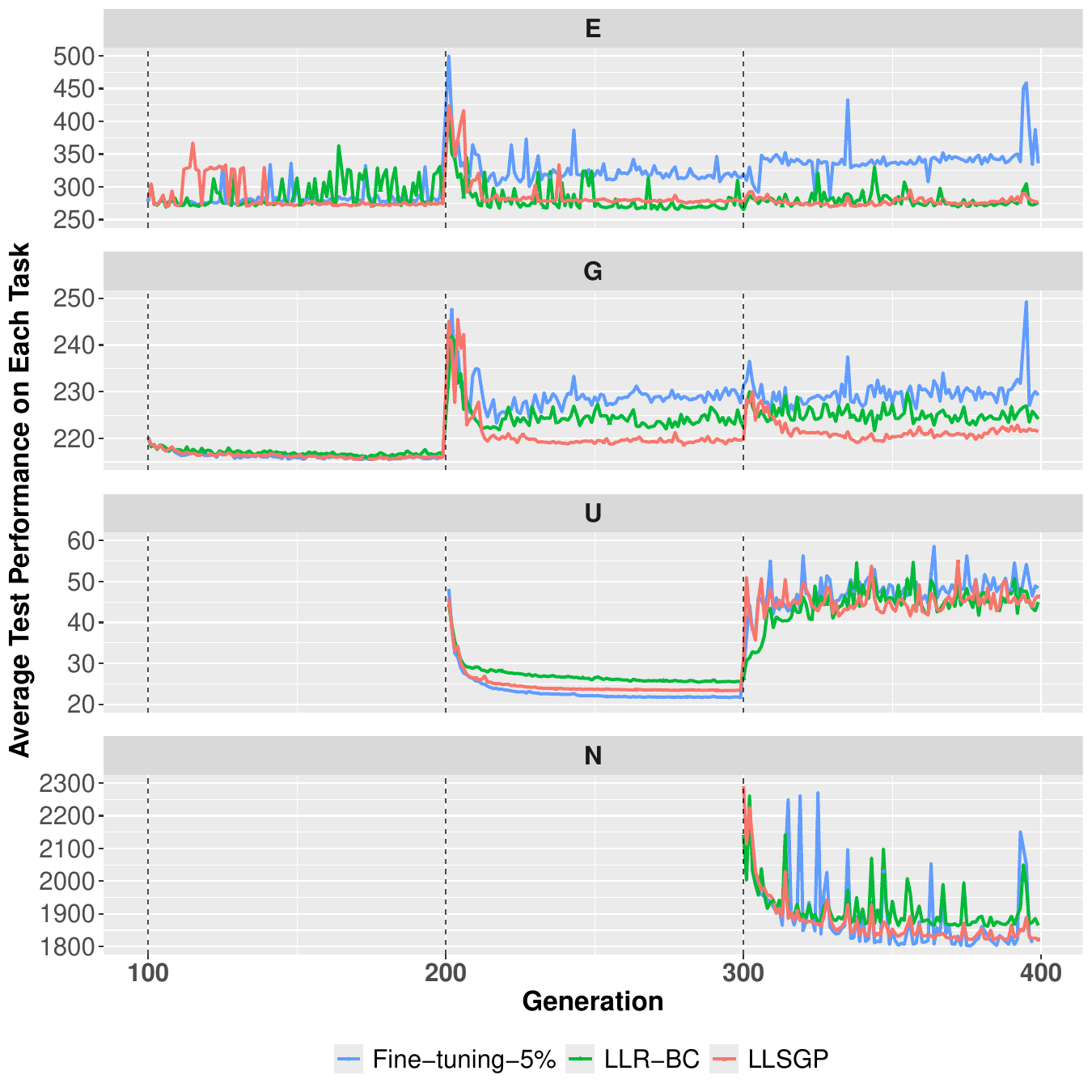}
	\vspace{-2mm}
	\caption{Average test performance on different tasks during lifelong learning under the EGUN training order.}
	\label{Fig. learning1}
\end{figure}

\begin{figure}[t]
	\centering 
	\includegraphics[width=0.47\textwidth]{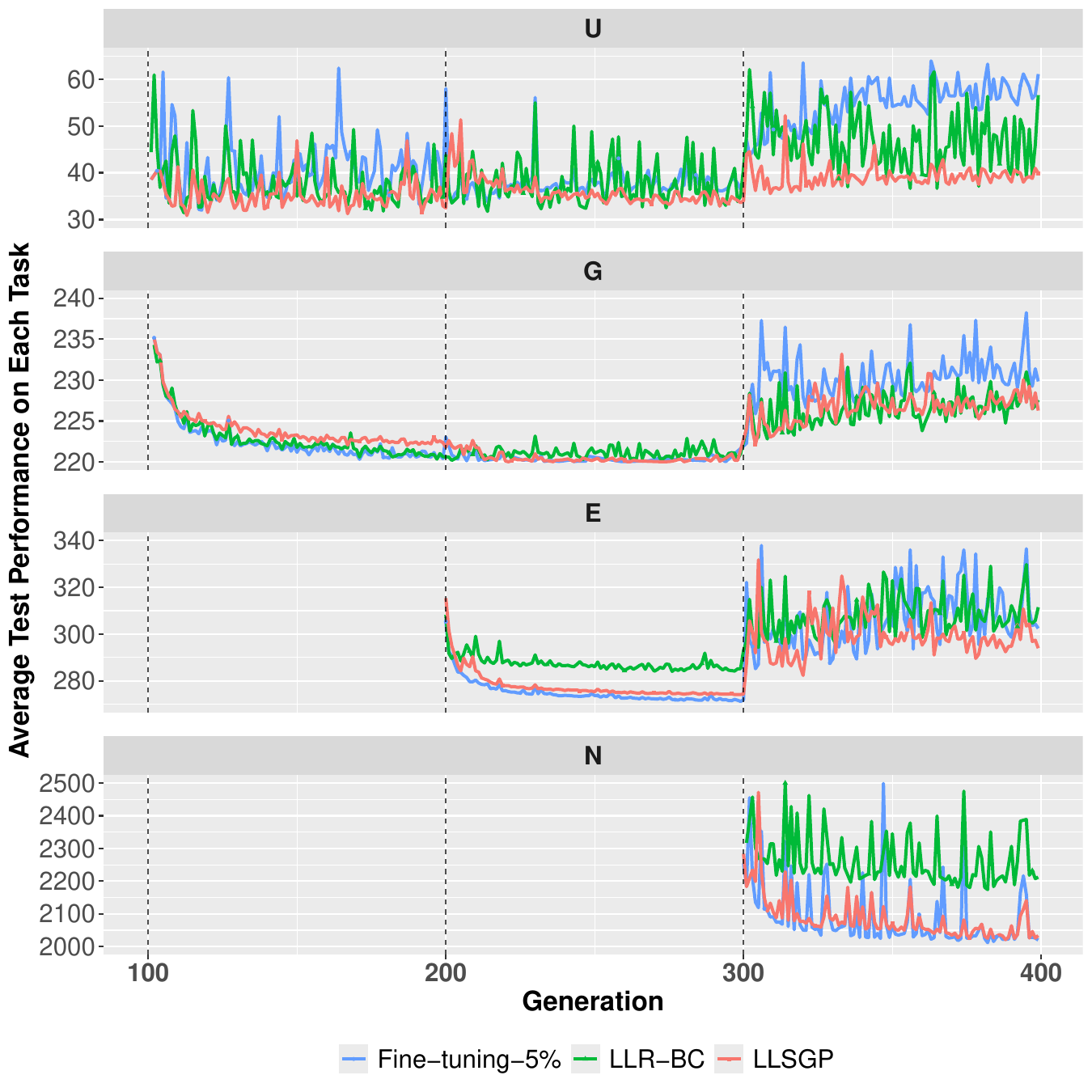}
	\vspace{-2mm}
	\caption{Average test performance on different tasks during lifelong learning under the UGEN training order.}
	\label{Fig. learning2}
\end{figure}

\begin{table*}[t]
  \centering
  \footnotesize
  \caption{Comparisons of mean (std) metrics for \textbf{ablation variations} over six training orders (30 runs each).}
  \label{tab:ablationResults}
  \resizebox{1\textwidth}{!}{
   \begin{tabular}{l@{\hspace{0.1pt}}l ccccc}
    \toprule
     \bfseries Metric && One-stage & Raw-surrogate & Equal-weight & Focused-weight & Proposed LLSGP  \\
    \midrule
    \multirow{2}{*}{AP}
      & Value
      & 44.38 (11.22) ($\uparrow$, 0.03)
      & 38.99 (7.43) ($\uparrow$, 0.04)
      & 36.11 (9.87) ($\approx$, 0.22)
      & 33.75 (9.41) ($\approx$, 0.22)
      & 29.75 (12.15) \\
      & Rank
      & 4.67 & 3.33 & 2.67 & 2.50 & 1.83 \\
    \midrule
    \multirow{2}{*}{AF}
      & Value
      & 35.52 (14.93) ($\approx$, 0.84)
      & 49.95 (12.75) ($\uparrow$, 0.04)
      & 39.15 (10.30) ($\approx$, 0.56)
      & 34.94 (14.43) ($\approx$, 0.84)
      & 34.35 (15.88) \\
      & Rank
      & 2.33 & 4.50 & 3.00 & 2.67 & 2.50 \\
    \midrule
    \multirow{2}{*}{AMF}
      & Value
      & 55.36 (28.90) ($\approx$, 0.84)
      & 92.41 (43.01) ($\uparrow$, 0.03)
      & 57.63 (12.07) ($\approx$, 0.69)
      & 54.36 (22.27) ($\approx$, 0.84)
      & 53.57 (27.70)\textbf{ } \\
      & Rank
      & 2.00 & 5.00 & 2.67 & 2.83 & 2.50 \\
    \midrule
    \multirow{2}{*}{APl}
      & Value
      & 10.61 (3.89) ($\uparrow$, 0.03)
      & 3.89 (0.67) ($\approx$, 0.31)
      & 5.55 (1.84) ($\approx$, 0.06)
      & 5.67 (1.54) ($\uparrow$, 0.03)
      & 4.12 (1.26) \\
      & Rank
      & 4.83 & 1.33 & 3.17 & 3.83 & 1.83 \\
    \midrule
    \multicolumn{2}{l}{Average Rank} & 3.46 & 3.54 & 2.88 & 2.96 & 2.17 \\
    \bottomrule
  \end{tabular}}
\end{table*}

\subsubsection{Stability}
To further analyse the stability of different algorithms, we compare the AF and AMF metrics reported in Table~\ref{tab:results}. Both metrics measure the degree of forgetting on previously learned tasks during lifelong learning. Because forgetting is measured as a relative gap, a low AF value alone does not necessarily imply better performance. Therefore, AF and AMF should be evaluated together with other metrics (e.g., AP and APl). LLSGP and LLR-BC achieve lower AF and AMF values than other compared methods. However, while LLR-BC exhibits low forgetting, this is primarily attributed to its highly limited plasticity, as it adapts less effectively to new tasks and therefore experiences smaller performance degradation. In contrast, LLSGP effectively retains past knowledge while maintaining strong plasticity on new tasks.

The learning curves in Fig.~\ref{Fig. learning1} and Fig.~\ref{Fig. learning2} provide further insight into the learning dynamics under two representative training orders. To improve the readability of the curves, only the best fine-tuning variant (Fine-tuning-5\%) is shown, and the first 100 generations are omitted because all methods exhibit almost identical performance. The dashed vertical lines indicate the switching points between tasks. After switching to a new task, the performance on previously learned tasks generally deteriorates as the search begins adapting to a new task distribution. For example, in Fig.~\ref{Fig. learning1}, when learning task $U$ at generation 200, the performance on tasks $E$ and $G$ changes abruptly due to the shift of optimisation focus. As training continues, both LLR-BC and LLSGP partially recover the performance on these tasks, indicating that they can mitigate forgetting to some extent. However, LLSGP consistently mitigates forgetting more effectively than LLR-BC, while Fine-tuning-5\% suffers the largest performance degradation. Similar patterns can also be observed in Fig.~\ref{Fig. learning2}. Moreover, LLSGP typically stabilises faster after switching to a new task, whereas other methods often experience larger fluctuations.


\subsubsection{Plasticity}
We further analyse the plasticity using the APl metric in Table~\ref{tab:results}. Restart and Fine-tuning methods achieve the good APl values, which is expected since their fitness evaluation depends only on the current task during evolution. LLSGP achieves a comparable APl value due to the proposed two-stage strategy. In the first stage, the search focuses entirely on the current task, enabling sufficient plasticity. This behaviour can also be observed in the learning curves. For example, in Fig.~\ref{Fig. learning1}, during the optimisation of task $U$ (generations 200--300) and task $N$ (generations 300--400), the performance of LLSGP on the current task is very similar to that of Fine-tuning-5\%. However, LLR-BC exhibits lower plasticity because incorporating behaviour similarity into the fitness function negatively affects performance on the current task. Overall, our proposed LLSGP can achieve better balance between plasticity and stability.

\subsection{Ablation Study}

To examine the contribution of the key components in the proposed LLSGP framework, we conduct an ablation study. Table~\ref{tab:ablationResults} shows the average results over six training orders. The full LLSGP achieves the best performance on all metrics, while removing or modifying any component leads to performance degradation.

\begin{enumerate}

\item \textit{Adaptive two-stage strategy}:  
The One-stage variant removes the adaptive switching mechanism and applies the Stage~2 fitness throughout the search. This leads to a clearly worse AP value (44.38) compared with LLSGP (29.75). In addition, the APl value increases noticeably (10.61 vs.\ 4.12), indicating weaker learning ability on the current task. Because it fails to learn the new task well, its performance cannot drop much later, which leads to a misleadingly low forgetting value. These results suggest that directly combining multi-task fitness throughout the search may hinder adaptation to new tasks. The adaptive two-stage strategy mitigates this issue by first focusing on the current task and then gradually incorporating knowledge from previously learned tasks.

\item \textit{QD-based surrogate with distance constraints}:  
The Raw-surrogate variant removes the proposed QD-based sample collection and the distance constraint. This leads to clear performance degradation in most metrics, with AP increasing from 29.75 to 38.99 and AF increasing from 34.35 to 49.95. Without diverse samples and distance filtering, the surrogate may provide unreliable fitness estimates for individuals far from existing samples, which can misguide the evolutionary search.

\item \textit{Task-weighting mechanism}:  
Two variants are used to examine the effect of task weighting. The \textit{Equal-weight} variant assigns identical weights to all tasks, while the \textit{Focused-weight} variant assigns weight 1 to the current task and distributes the remaining weight equally among previous tasks. Both variants achieve worse plasticity ($APl = 5.55$ and $5.67$ vs. $4.12$) and lower overall ranks ($2.88$ and $2.96$ vs. $2.17$). Although the difference in individual metrics is not always statistically significant, the proposed LLSGP consistently obtains the best mean values and the top overall rank across all metrics.

\end{enumerate}

Overall, the ablation results show that all three components contribute to the effectiveness of LLSGP. The adaptive two-stage strategy improves the balance between plasticity and stability, the QD-based surrogate enhances the reliability of surrogate-assisted selection, and the task-weighting mechanism assigns larger weights to less similar tasks, helping reduce uneven forgetting and improve overall performance.


\subsection{Analysis of Task Weights}

\begin{figure}[t]
	\centering 
	\includegraphics[width= 0.5 \textwidth]{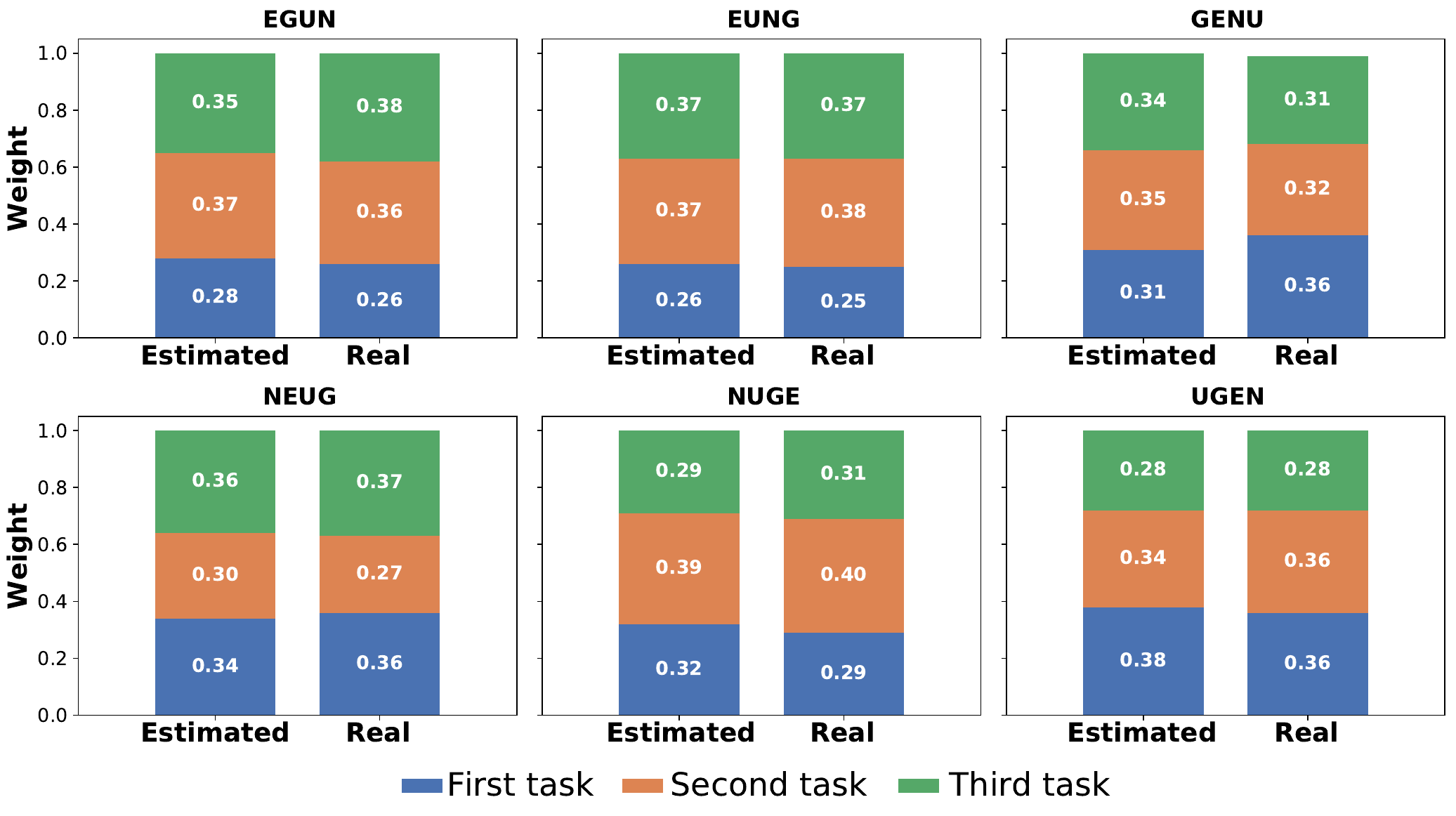}
	\vspace{-2mm}
	\caption{Average weights of tasks across six training orders.}
    \label{fig:task_weight_comparison}
\end{figure}

To calculate task similarity weights without requiring expensive simulation evaluations, the proposed method estimates task correlations based on the performance rankings of representative individuals. To verify the reliability of these surrogate-estimated weights, we compared the estimated task weights ($W_{\text{est}}$) and the ground-truth weights ($W_{\text{real}}$, computed using exact simulations) across all six training sequences, as illustrated in Fig.~\ref{fig:task_weight_comparison}.

In our framework, the retention weights assigned to historical tasks are normalised such that $\sum_{k=1}^{t-1} w_k = 1$. Each stacked bar in Fig.~\ref{fig:task_weight_comparison} represents the relative retention weights assigned to the first, second, and third tasks when training on the fourth task. Across all six training orders, $W_{\text{est}}$ closely aligns with $W_{\text{real}}$ with only negligible deviations (e.g., in the EGUN sequence, the estimated weights of $0.28$, $0.37$, and $0.35$ closely track the true weights of $0.26$, $0.36$, and $0.38$). This empirical consistency confirms that lightweight surrogate estimation provides an accurate and reliable estimation for true task relationships without incurring additional simulation overhead.

Furthermore, across different training orders, the weights vary due to different task similarity relationships. In general, the proposed strategy assigns larger weights to less similar tasks and smaller weights to more similar ones. This helps mitigate forgetting while preventing the search from relying too heavily on any single previous task.

\begin{table}[t]
  \centering
  \footnotesize
  \caption{Average generation at which the algorithm switches from Stage~1 to Stage~2 across six training orders.}
  \label{tab:stage_switch}
  \resizebox{0.30\textwidth}{!}{
  \begin{tabular}{ccc}
    \toprule
    Task 2 & Task 3 & Task 4 \\
    \midrule
    107.5 (2.0) & 207.3 (1.6) & 306.8 (1.1) \\
    \bottomrule
  \end{tabular}}
\end{table}

\subsection{Stage Switching Behaviour}

Table~\ref{tab:stage_switch} shows the average generation at which the algorithm switches from Stage~1 to Stage~2 across different tasks. The switching points occur around generations 108, 207, and 307 when learning Tasks~2, 3, and 4, respectively.

The small standard deviations indicate that the switching behaviour is highly consistent across training orders. In practice, the algorithm performs Stage~1 optimisation in the early phase of each task and switches to Stage~2 once the improvement in population performance becomes limited. This pattern is also reflected in Fig.~\ref{Fig. learning1} and Fig.~\ref{Fig. learning2}, where the performance on the current task typically stabilises within about ten generations before the switching point. These results suggest that the proposed population improvement based switching mechanism can reliably determine when to incorporate knowledge from previously learned tasks, helping balance plasticity and stability during lifelong learning.

\begin{figure}[t]
	\centering 
	\includegraphics[width= 0.48 \textwidth]{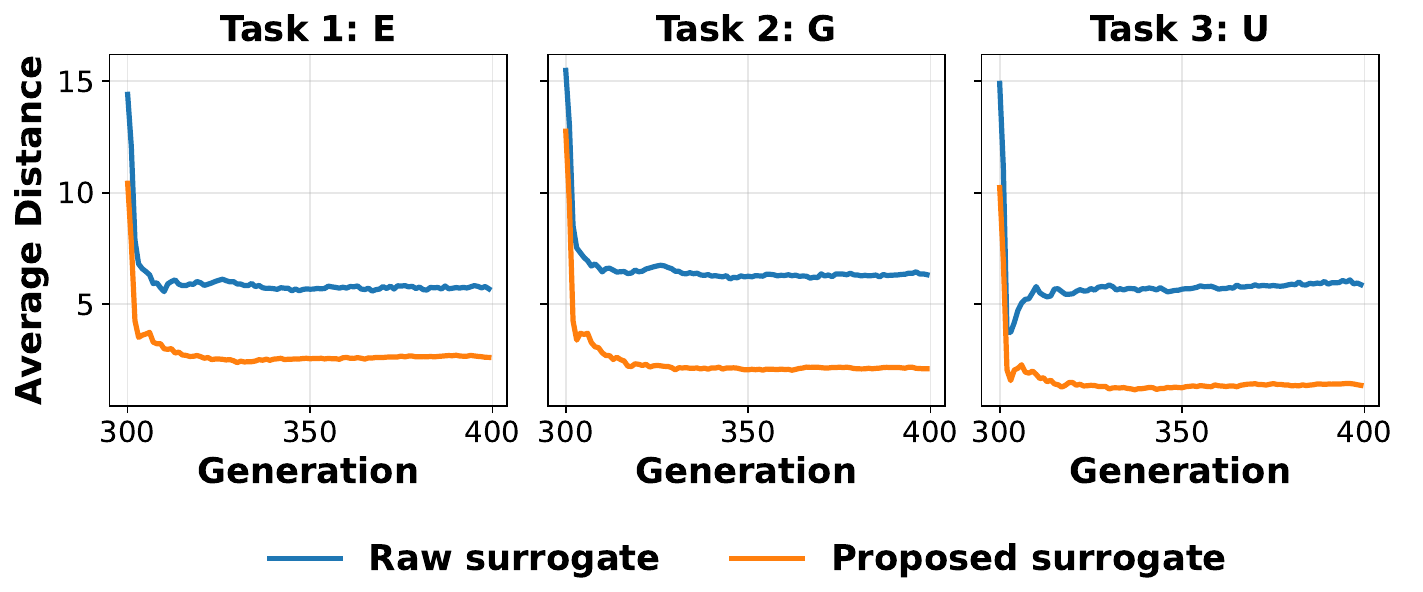}
	\vspace{-2mm}
	\caption{Average behaviour vector distances to surrogate samples of the three previously learned tasks when learning the fourth task (EGUN order).}\label{Fig. distance}
\end{figure}

\begin{figure}[t]
	\centering 
	\includegraphics[width= 0.48 \textwidth]{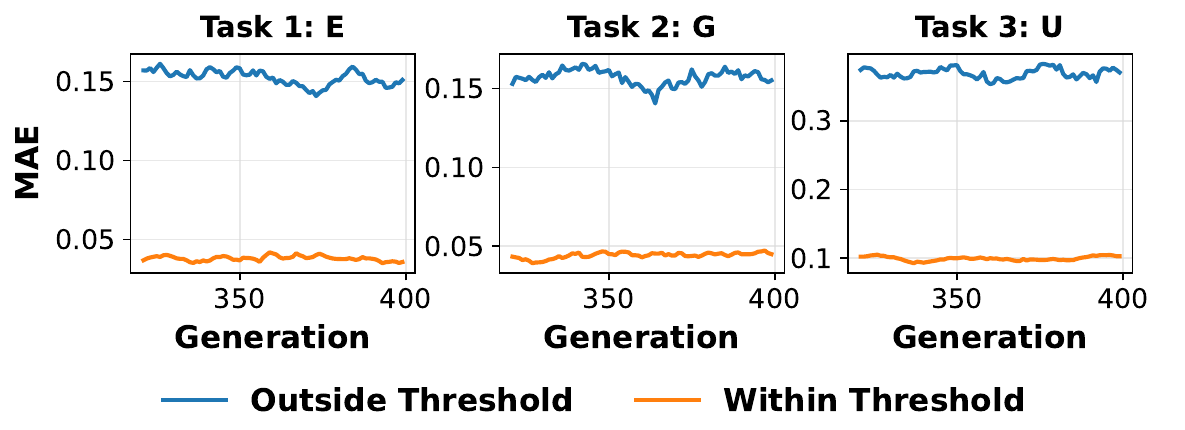}
	\vspace{-2mm}
	\caption{MAE comparison between the regions within and outside the threshold across the three previously learned tasks during the training of the fourth task under the EGUN order}\label{Fig. MAE}
\end{figure}

\begin{table*}[t]
  \centering
  \footnotesize
  \caption{Mean (std) training time (in minutes) of the algorithms across six training orders}
  \label{tab:training time}
  \resizebox{0.84\textwidth}{!}{
  \begin{tabular}{llllll}
    \toprule
     Restart & Fine-tuning-5\% & Fine-tuning-10\% & Fine-tuning-100\% & LLR-BC & Proposed LLSGP \\
\midrule
383.09 (60.36) & 458.16 (89.93) & 461.24 (90.10) & 479.24 (92.61) & 461.72 (89.12) & 485.44 (99.85) \\
    \bottomrule
  \end{tabular}}
\end{table*}

\subsection{Surrogate Accuracy}

To investigate whether the proposed surrogate improves estimation accuracy, we analyse the behaviour vector distances between individuals in the current population and the surrogate samples collected from previously learned tasks. Following \cite{DBLP:journals/ec/HildebrandtB15}, the accuracy of surrogate estimation depends on the distance between the decision-making vector of an evaluated individual and its nearest surrogate sample: smaller distances generally indicate more reliable fitness estimates.

Fig.~\ref{Fig. distance} illustrates the average distances to the surrogate samples of the three previously learned tasks when learning the fourth task under the EGUN training order. Similar patterns are observed for the other training orders. Overall, the proposed surrogate consistently maintains smaller distances than the classical surrogate. The distances are initially large because newly initialised individuals are not yet aligned with the historical behaviour regions represented by the surrogate samples. As evolution proceeds, the distances decrease rapidly and stabilise at a lower level. This indicates that the proposed surrogate keeps the population closer to existing surrogate samples, leading to more reliable surrogate-assisted evaluations during lifelong learning.



In addition, to further verify the prediction accuracy of the proposed surrogate model with a distance threshold, we investigate whether the predictions for individuals within the threshold are more accurate than those outside it. To achieve this, we employ the Mean Absolute Error (MAE) as the evaluation metric, which directly measures the absolute deviation between the surrogate-estimated fitness and the true simulator fitness, following the normalisation method described previously. Crucially, to illustrate the potential estimation risk, the MAE for the outside-threshold region is calculated using the baseline nearest-neighbour estimation instead of an infinite fitness. Fig.~\ref{Fig. MAE} illustrates the MAE comparison between the regions within and outside the threshold across the three previously learned tasks during the training of the fourth task under the EGUN order. Clearly, among all three tasks, the MAE within the threshold is significantly lower than that outside it. This demonstrates that the distance threshold successfully identifies the high-fidelity region, thereby improving the reliability of the surrogate model.


\section{Further Analyses}

\subsection{Training time}

To investigate whether the proposed LLSGP introduces much extra computational cost, Table~\ref{tab:training time} presents the mean (std) total training time (in minutes) of all compared algorithms across the six training orders. The results indicate that the total training time of the proposed LLSGP 
is slightly higher than that of the other algorithms. This marginal increase is primarily attributed to the surrogate construction process executed after completing the training of each individual task. Note that the total training time reported for LLSGP already includes all computational steps: the evolutionary process of traditional GP, the surrogate construction 
, and the surrogate estimations during preselection. 

\subsection{Sizes of the Best Learned Scheduling Rule Pair}

\begin{figure}[t]
	\centering 
	\includegraphics[width= 0.48 \textwidth]{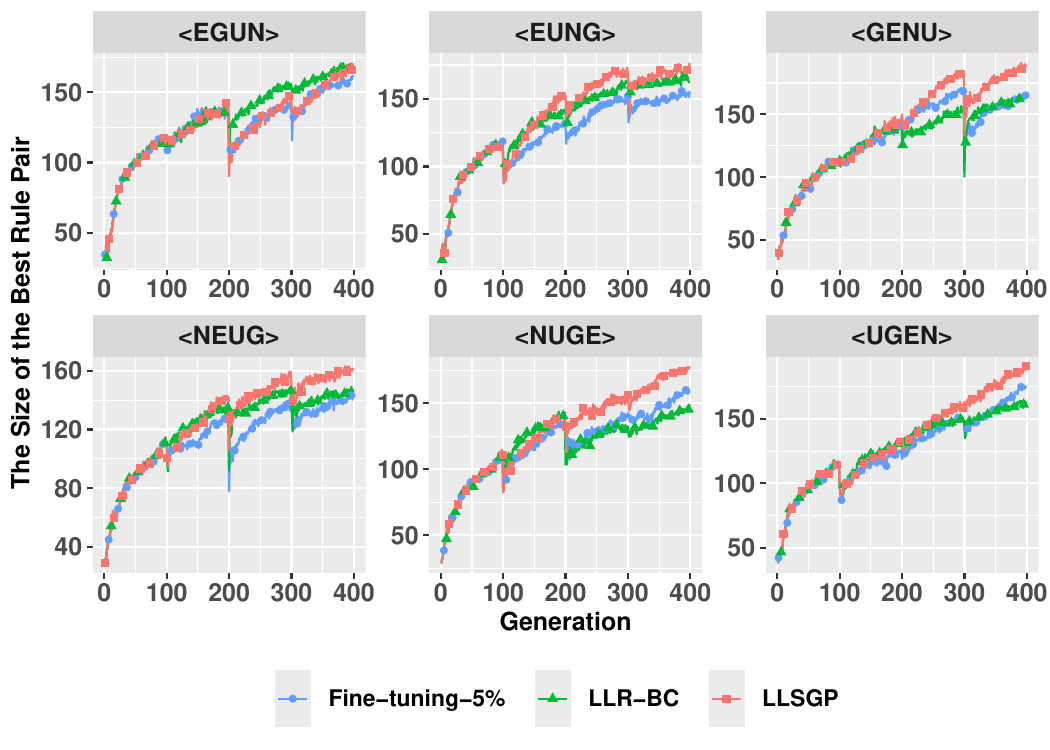}
	\vspace{-2mm}
	\caption{Size of best evolved scheduling rule pair across six training orders during evolutionary process} \label{Fig. ruleSize}
\end{figure}

We further analyse the sizes of the best learned rule pair during lifelong learning. Fig.~\ref{Fig. ruleSize} shows the size (total number of nodes) of the best-evolved scheduling rule pair across the six training sequences throughout the evolutionary process. Overall, the rule size increases steadily over generations, indicating that more complex rule structures are evolved to handle multiple tasks. Noticeable drops occur around the task-switching points (approximately generations 100, 200, and 300), as randomly generated individuals with smaller rule sizes are introduced when a new task arrives. After each switch, the rule size quickly grows again as the population adapts to the new task. Moreover, the growth rate differs among algorithms: fine-tuning increases the slowest, LLR-BC grows faster, while the proposed LLSGP shows the fastest growth and generally achieves the largest rule sizes in later generations. A possible reason for this size increase is a more effective exploration and knowledge accumulation across tasks.

\section{Conclusions}

The goal of this paper was to develop a lifelong learning method that enables GP to handle sequential heterogeneous DFJSS tasks with job/batch arrivals. This goal was successfully achieved by proposing a surrogate-assisted GP framework to balance plasticity and stability during lifelong learning.

Experimental results demonstrate that the proposed method outperforms the compared algorithms across multiple lifelong training orders. This superior performance stems from its ability to effectively balance plasticity and stability, enabling rapid adaptation to new tasks while retaining knowledge from previously learned ones. By leveraging task similarity to regulate knowledge reuse, the framework promotes the evolution of more generalisable scheduling rules. Furthermore, the integration of distance-aware surrogate evaluation enhances the reliability of fitness estimation, leading to more robust evolutionary search in lifelong learning scenarios.


Future work will investigate more complex scheduling environments, such as boundary-free scenarios and extended task sequences. Furthermore, we plan to explore knowledge reuse at the sub-tree level to further mitigate catastrophic forgetting.

%

\ifCLASSOPTIONcaptionsoff
  \newpage
\fi

%
%

\bibliographystyle{IEEEtran}
\bibliography{mybib} 

\end{document}